%% file: ICANN2018.tex
\documentclass[runningheads]{llncs}

\usepackage{amssymb}
\usepackage{graphicx}
\usepackage{url}
\usepackage{subfig}
\usepackage{multirow}
\usepackage{booktabs}
\usepackage{array}
\usepackage{enumitem}
\usepackage[labelfont=bf]{caption}
\newcolumntype{L}{>{$}l<{$}}
\newcolumntype{C}{>{$}c<{$}}
\newcolumntype{R}{>{$}r<{$}}

\usepackage{url}
\urldef{\mailsa}\path|tiezzi24@student.unisi.it|
\urldef{\mailsb}\path|mela@diism.unisi.it|
\urldef{\mailsc}\path|maggini@diism.unisi.it| 
\urldef{\mailsd}\path|a.frosini@istech.it|
\newcommand{\keywords}[1]{\par\addvspace\baselineskip
\noindent\keywordname\enspace\ignorespaces#1}

\usepackage{url}

\begin{document}

\mainmatter  % start of an individual contribution

\title{Video Surveillance of Highway Traffic Events by Deep Learning Architectures  \thanks{This is a post-peer-review, pre-copyedit version of an article published in
LNCS, volume 11141. The final authenticated version is available online at: \protect\url{https://doi.org/10.1007/978-3-030-01424-7_57}}}

\titlerunning{Video Surveillance of Highway Traffic Events by Deep Learning   \thanks{This is a post-peer-review, pre-copyedit version of an article published in
LNCS, volume 11141. The final authenticated version is available online at: \protect\url{https://doi.org/10.1007/978-3-030-01424-7_57}}}

\author{Matteo Tiezzi \inst{1}\and Stefano Melacci \inst{1}\and Marco Maggini \inst{1}\and Angelo Frosini \inst{2}}
\authorrunning{Tiezzi, Melacci, Maggini, Frosini}
%\titlerunning{Video Surveillance by Deep Learning}

\institute{
Department of Information Engineering and Mathematics, University of Siena \\
\url{http://sailab.diism.unisi.it} \\
\email{\{mtiezzi,mela,maggini\}@diism.unisi.it}
\and
	IsTech s.r.l., Pistoia, Italy\\ 
\email{a.frosini@istech.it}
}


%\toctitle{Character context based neural model}
%\tocauthor{Authors' Instructions}
\maketitle

\begin{abstract}
In this paper we describe a video surveillance system able to detect traffic events in videos acquired by fixed videocameras on highways.
The events of interest consist in a specific sequence of situations that occur in the video, as for instance a vehicle stopping on the
emergency lane. Hence, the detection of these events requires to analyze a temporal sequence in the video stream.
We compare different approaches that exploit architectures based on Recurrent Neural Networks (RNNs) and Convolutional Neural Networks (CNNs).  
A first approach extracts vectors of features, mostly related to motion, from each video frame and exploits a RNN fed with the
resulting sequence of vectors. The other approaches are based directly on the sequence of frames, that are eventually enriched with pixel-wise motion
information. The obtained stream is processed by an architecture that stacks a CNN and a RNN, and we also investigate a transfer-learning-based model. The results are very promising and the best architecture
will be tested online in real operative conditions.

\keywords{Convolutional Neural Networks, Recurrent Neural Networks, Deep Learning, Video Surveillance, Highway Traffic}
\end{abstract}

\section{Introduction}
\label{sec:intro}
\input{introduction.tex}

\section{Video Surveillance of Highway Traffic} % PROBLEM
% problema di riconoscere certi eventi (quali) in segnali video di telecamere fisse che sorvegliano autostrade
% foto di esempio (screenshot) PNG dei vari eventi
% elenco classi
% difficolt tipiche che si hanno nel cercare di fare predizioni in questi video (luminosit-notte, disturbi, situazioni anomale, eventi climatici, ...)
\label{sec:problem}
\input{problem.tex}

\section{Data Description and Representation} % DATA
% dati forniti da isTech (???) o da una "certa" azienda
% caratteristiche intrinseche (risoluzione, framerate, ...)
% (i dati sono un campione limitato)
% (differenti preprocessing suggeriti dall'azienda per ulteriori indagini)
% tabella con statistiche esatte sui dati (esempi per classe, lunghezza, ...)
% descrivere i preprocessing (optical flow, istogrammi, ...)
\label{sec:data}
\input{data.tex}

\section{Deep Architectures} % MODELS
% funzione di costo (pesatura da descrivere)
% descrivere i modelli neurali valutati (2/3, anche la rete con la parte preaddestrata)
% figura architetture
% qualche formula azzardata per descrivere i modelli
\label{sec:model}
\input{model.tex}

\section{Experimental Results} % RESULTS
% setup sperimentale (training, validation, test set, misura di accuratezza usata)
% tempi di addestramento in una macchina di riferimento (1 macchina)
% risultati (tabella senza VGG), 1 e 2 layers di RNNs
% discussione
% risultati VGG (tabella aggiuntiva)
% discussione
% prendere una classe che funziona peggio delle altre, prendere un esempio di quella classe, mostrare il grafico delle predizioni e della ground truth, commentare (anche uno, due frame di esempio)
\label{sec:experiments}
\input{experiments.tex}

\vskip-5mm
\section{Conclusions}
\label{sec:conclusions}
\input{conclusions.tex}
\vskip-3mm
% EITHER use the included BST file
\bibliographystyle{splncs04}
\bibliography{ICANN2018}

\end{document}

%% file: introduction.tex
% !TEX root = icann2018.tex

The progressive growth of the number of vehicles, that nowadays are traveling on roads and highways, has created high interest in the research areas
related to the development of techniques needed in automatic instruments for traffic monitoring. These systems are generically
referred to as Intelligent Transportation Systems (ITSs). Basic tasks, that are to be accomplished by ITSs, are the identification
of vehicles and of their behaviour from video streams, captured by surveillance cameras installed along
the road connections. The automatic detection of specific events happening in the traffic flow, such as accidents, dangerous driving, and
traffic congestions, has become an indispensable functionality of ITSs since it is impractical to employ human operators
both for the number of control points and the need of a continuous attention. Automatic notifications guarantee 
an immediate response to exceptional events such as car crashes or wrong--way driving. At the same time, the estimation of road congestion
allows us to notify drivers and to provide information for optimizing the itineraries computed by navigation devices. 
This field of research began to be particularly active in the `80, with projects funded by governments, industries and universities, in Europe (PROMETHEUS \cite{Pro}), Japan (RACS \cite{racs}) and the USA (IVHS \cite{ivhs}). These studies included autonomous cars, inter--vehicle communication systems \cite{lee2010collaborative}, surveillance and monitoring of traffic events \cite{michalopoulos1989vehicle,coifman1998real}. 

Among the general ITSs, the Advanced Traffic Management Systems (ATMS) are aimed at exploiting all the information coming from cameras, sensors and other instruments, positioned along highways and main routes, to provide an analysis of the current state of traffic and to respond in real time to specific conditions. Signals from all 
devices are gathered at a central Transportation Management Center that must implement technologies capable of analyzing the huge amounts of data coming from all the sensors and cameras. 

In this context, Machine Learning provides tools to tackle many problems faced in the design of the ATMS modules.
In particular, Deep Neural Network architectures are able to yield state-of-the-art performances in many computer vision tasks \cite{he2016deep} and
are currently applied in real systems, such those for autonomous driving \cite{chen2015deepdriving}. Hence, most of current video surveillance modules are based on deep
learning techniques, that allow us to tune the system just by providing enough examples of the objects or events of interest \cite{XU2017117}.
The wide use of these approaches has also be driven by the availability of pre--trained architectures for computer vision
tasks that can be adapted to new problems by transfer learning \cite{oquab2014learning}. 

The objective of this work is the creation of an instrument capable to perform a real time/on-line analysis of data coming from 
cameras, in order to detect automatically significant events occurring in the traffic flow. We analyze the results obtained by
different approaches on real videos of traffic on highways. In particular we compare an approach based on precomputed
motion features processed by a Recurrent Neural Network (RNN) with a technique exploiting the original video augmented by
channels to encode the optical flow. The latter is based on an architecture composed by a Convolutional Neural Network (CNN), processing each
input frame, stacked with a RNN. We consider both the cases in which the CNN is learned from our traffic videos and when it is a pre--trained CNN in a transfer learning scheme.

The paper is organized as follows. The next Section describes the considered problem, while our dataset and the feature representation are described in Section \ref{sec:problem}. In Section \ref{sec:model} we introduce the selected deep neural network architectures, while Section \ref{sec:experiments} reports the results. Finally, Section \ref{sec:conclusions} concludes the paper.

%% file: problem.tex
% !TEX root = icann2018.tex

\begin{figure}[!ht]
  \centering{
  \subfloat[a][\label{fermo}]{\includegraphics[width=0.24\columnwidth]{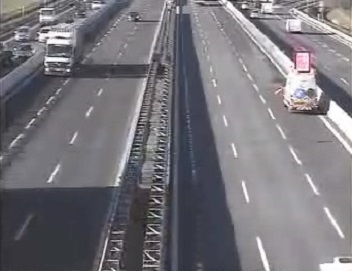} }
  \subfloat[b][\label{rip}]{\includegraphics[width=0.24\columnwidth]{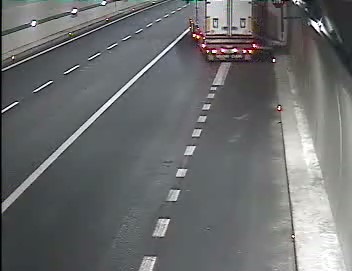} }
  \subfloat[c][\label{contro}]{\includegraphics[width=0.24\columnwidth]{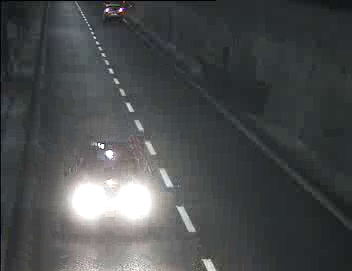} }
  \subfloat[d][\label{inci}]{\includegraphics[width=0.24\columnwidth]{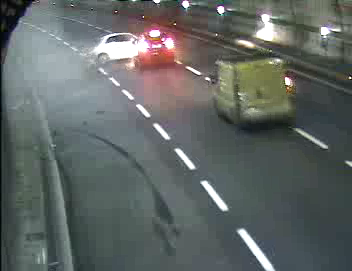} }
  \caption{\label{fig:classes} The four different classes of events. \protect\subref{fermo} Stationary.  \protect\subref{rip} Departing.  \protect\subref{contro} Wrong-way. \protect\subref{inci} Car crash.}
}
\end{figure}

We focus on a system that processes videos acquired by fixed cameras on highways. Cameras can be positioned in very
different environments (e.g. tunnels or outdoor) and can have many different settings for the point of view (e.g. long
or short range, wide or narrow span). Moreover, videos are captured in different environmental and weather conditions (daylight, night,
fog, rain, etc.). The system is expected to detect specific events of interest happening in the scene for a variable time interval. In particular, we consider four different classes
of events, collected in the set $\mathcal{E}$ (see Fig.~\ref{fig:classes}):
\begin{itemize}[leftmargin=*]
\item\hskip-0.7mm  \textbf{Stationary vehicle}, a vehicle stops inside the field of the camera;
\item\hskip-0.7mm  \textbf{Departing vehicle},  a vehicle, previously stationary, departs from his position;
\item\hskip-0.7mm  \textbf{Wrong--way vehicle}, a vehicle moves in the wrong direction;
\item\hskip-0.7mm \textbf{Car crash}, accident involving one or more vehicles.
\end{itemize}

%\begin{figure}[h]
%\includegraphics[width=0.24\columnwidth]{images/fermo_07_Moment} 
%\includegraphics[width=0.24\columnwidth]{images/riparte} 
%\includegraphics[width=0.24\columnwidth]{images/contro} 
%\includegraphics[width=0.24\columnwidth]{images/inci} 
%\centering
%\caption{The four different classes of events, Stationary , Departing and Wrong-way vehicle, Car Crash}
%\centering
%\label{figura_classi}
%\end{figure}

\begin{figure}[!ht]
  \centering{
  \subfloat[a][\label{rain}]{\includegraphics[width=0.24\columnwidth]{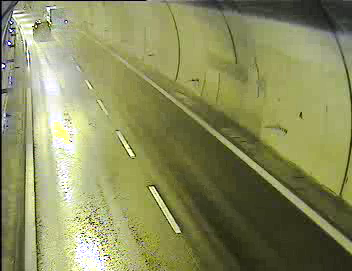} }
  \subfloat[b][\label{fog}]{\includegraphics[width=0.24\columnwidth]{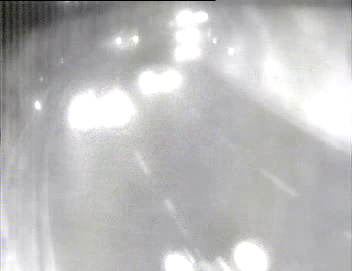} }
  \subfloat[c][\label{before}]{\includegraphics[width=0.24\columnwidth]{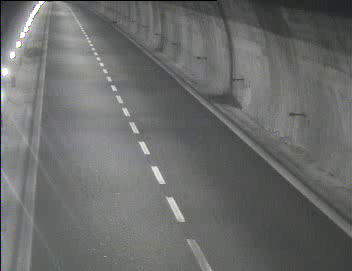} }
  \subfloat[d][\label{after}]{\includegraphics[width=0.24\columnwidth]{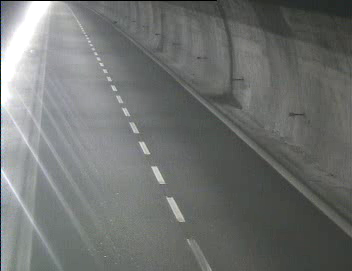} }
  \caption{\label{fig:conditions} Conditions causing difficulties in video analysis.  \protect\subref{rain} Rain.  \protect\subref{fog} Fog.  \protect\subref{before}-\protect\subref{after} Brightness variation, before and after.}
}
\end{figure}

As already stated, all the videos are captured by cameras positioned in different places and settings on highways, including tunnels and high--speed stretches.
This fact entails several issues that can deteriorate the prediction performances.
For instance, cameras are exposed to all kind of weather conditions, including fog, rain or strong wind.
Another relevant problem is due to variations in brightness caused by tunnel lamps activation, clouds passing by, and sun movement (see Fig.~\ref{fig:conditions}
for examples).

%\begin{figure}[h]
%\includegraphics[width=0.24\columnwidth]{images/rain} 
%\includegraphics[width=0.24\columnwidth]{images/nebbia} 
%\includegraphics[width=0.24\columnwidth]{images/luce2} 
%\includegraphics[width=0.24\columnwidth]{images/luce1} 
%\centering
%\caption{Particular tipical conditions causing difficulties in video analysis}
%\centering
%\label{figura_classi}
%\end{figure}

%% file: data.tex
% !TEX root = icann2018.tex

Video surveillance cameras provide a continuous stream of a given view of the highway along the direction of the traffic flow.
Due to the nature of the events we are trying to detect, it was difficult to collect a large dataset of examples\footnote{The dataset was collected thanks to IsTech srl
and was based only on a limited number of fixed cameras.}. For instance, some events like wrong--way driving are quite rare.

Videos were captured in colors in two standard resolutions ($352\times288$ and $640\times320$ pixels, depending on the camera type) at $25$ frames per second.
In some cases the videocamera includes the lanes in both directions in its field. Hence, in order to remove potential sources of misleading information
(for instance, related to wrong--way vehicles) each frame is masked with a template that keeps only the portion related to the lanes to be
considered (see Fig. \ref{fig:masking}).
\begin{figure}[!ht]
  \centering{
  \subfloat[a][\label{frame}]{\includegraphics[width=0.24\columnwidth]{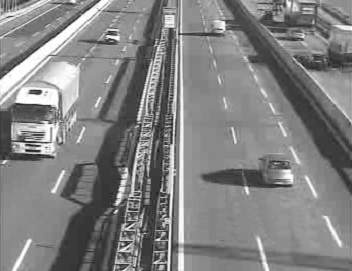}\ }
  \subfloat[b][\label{mask}]{\ \includegraphics[width=0.24\columnwidth]{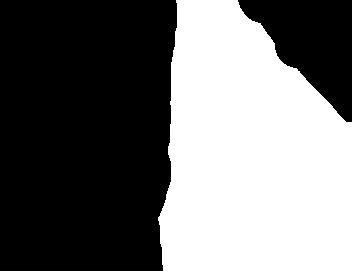}\ }
  \subfloat[c][\label{masked_frame}]{\ \includegraphics[width=0.24\columnwidth]{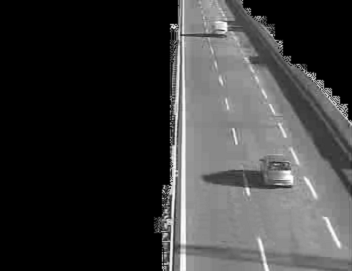}}
  \caption{\label{fig:masking} \protect\subref{frame} Original frame. \protect\subref{mask} Mask. \protect\subref{masked_frame} Masked frame.}
}
\end{figure}

We down--sampled the available videos at 2 frames per second, and  extracted clips of $125$ frames (1 minute), containing instances of the events $\mathcal{E}$ listed in Section \ref{sec:problem}, as well as clips with normal traffic conditions. To avoid artificial regularities that may hinder the generalization,
the clips are generated such that events can happen in every instant inside the $125$ frame interval, apart from the
very beginning or ending. The statistics of the available dataset used in training and testing are reported in Tab. \ref{tab:data2}.
%\begin{table}[]
%\centering
%\label{tab:data}
%\begin{tabular}{@{}lll@{}}
%\toprule
%Event  class &   \# Clips \\ \midrule
%No event  &      281        \\
%Stationary &     56       \\
%Departing &     111        \\
%Wrong-way &     131         \\ 
%Car crash &     16       \\ \midrule
%Total &     595  \\ \bottomrule
%\end{tabular}
%\caption{\label{tab:data} Statistics of the dataset used in the experiments.}
%\end{table}
\begin{table}[]
\centering
\label{tab:data}
\caption{\label{tab:data2} Statistics of the dataset used in the paper.}
\vskip 1mm
\begin{tabular}{l|ccccc|c}
\hline
\hline
 & $\ $No Event$\ $ & $\ $Stationary$\ $ & $\ $Departing$\ $ & $\ $Wrong-way$\ $ & $\ $Car crash$\ $ &  $\ $\textbf{Total}$\ $ \\ \hline
$\ $\textbf{\# Clips}$\ $  &      281 & 111 & 56 & 131 & 16 & 595 \\ \hline\hline
\end{tabular}
\end{table}
%\begin{table}
%\centerline{
%\begin{tabular}{|c|c|c|}
%\hline 
%\textbf{No event}   && 281   \\ \hline 
%\textbf{Stationary}  && 56 \\ \hline
%\textbf{Departing}  && 111 \\ \hline
%\textbf{Wrong-way}  && 131 \\ \hline
%\textbf{Crash}  && 16 \\ \hline
%\textbf{Total}  && 595 \\ \hline
%\end{tabular}}
%\caption{\label{tab:data} Statistics of the dataset used in the experiments.}
%\end{table}
%\begin{table}
%  \centering
%  \begin{tabular}
%
%No event & 281   \\ 
%Stationary & 2.85  \\ 
%Departing & 2.85  \\ 
%Wrong-way & 4.52\\ 
%Crash & 4.52  \\ 
%
%
%\end{tabular}
%\end{table}
The optical flow algorithm\footnote{We used the default implementation in the OpenCV library \url{https://opencv.org/}, based on the Farneback's algorithm.} was exploited to compute the motion field for each input frame. Each frame was 
resized and cropped to $160\times120$ pixels.
We represented the input frames in three different ways, using $i.$ pre-designed motion features, $ii.$ appearance, or $iii.$ appearance and motion, as described in the following.
$\ $\\
\vskip -2mm

\begin{figure}[!ht]
  \centering{
  \subfloat[a][\label{fig:optf}]{\includegraphics[width=0.249\columnwidth]{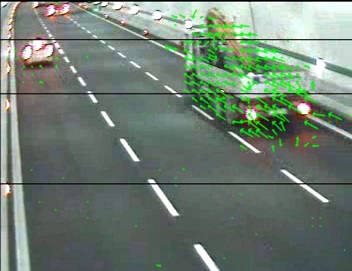}\ }
  \subfloat[b][\label{fig:histo}]{\includegraphics[width=0.23\columnwidth]{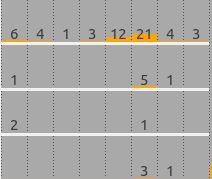}\ }
  \subfloat[c][\label{fig:backg}]{\includegraphics[width=0.505\columnwidth]{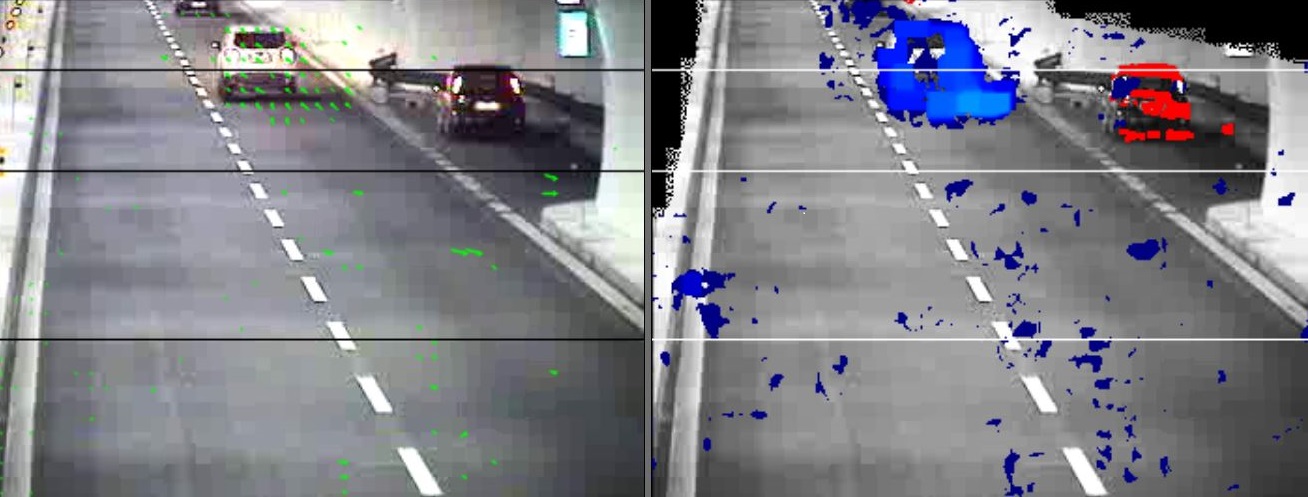}}
  \caption{\label{fig:masking} \protect\subref{fig:optf} Motion vectors, frame partitioned into 4 stripes. \protect\subref{fig:histo} Histogram computed in one stripe. \protect\subref{fig:backg} Background subtraction (original frame on the left, estimated not-background objects on the right).}}
\end{figure}

\noindent\textbf{Representation by motion features.} 
Due to the effect of perspective, moving objects closer to the camera position have an apparent motion larger than distant objects.
Therefore, we decided to split each frame into four horizontal stripes as shown in Fig. \ref{fig:optf}.
For each stripe the directions and modules of the optical flow are quantized, building a histogram of the distribution of the motion vectors. In the implementation we considered
32 bins  based on 8 directions and 4 levels for the module (Figure \ref{fig:histo}). This scheme yields $128$ values ($32$ bins for each stripe) collected into a vector
for each frame. In order to provide evidence for stationary vehicles, we computed an additional feature for each stripe as follows. We applied and manually tuned a Background Subtraction \cite{kaewtrakulpong2002improved} method to extract the pixels not belonging to the static background of the video (see Figure \ref{fig:backg}). The additional feature per stripe is the count of non--background pixels having null motion. %{\bf The background estimation algorithm is configured to provide a slow update and it is suspended when a stationary vehicle is detected.} 
Hence, each frame is represented by a vector of $132$ entries.
$\ $\\
\vskip -2mm

\noindent\textbf{Representations by appearance and motion.}
The appearance-based representation consists of the raw frame converted to grayscale to reduce the image variability. %A first representation is a video with a resolution $160\times120$ pixels and one channel for the gray level. 
Another representation is obtained by adding two additional channels for each frame corresponding to the horizontal and vertical components
of the motion field provided by the optical flow, leading to a $160\times120\times3$ tensor. 
%\begin{figure}[!ht]
%  \centering{
%{\includegraphics[width=0.2\columnwidth]{images/opt}}
%  \caption{\label{fig:histogram} Motion features for each frame.}
%}
%\end{figure}

%% file: model.tex
% !TEX root = icann2018.tex

% INFERENCE OF THE GENERIC NET (NO STRUCTURE)

We are given a video stream $\mathcal{V}$ that produces frames at each time instant $t$. At a certain $t>0$, we have access to the sequence of frames up to time $t$, that we indicate with $\mathcal{S}_t =  \{ \mathcal{I}_i,\ i=1,\ldots,t \}$, where $\mathcal{I}_i$ is the $i$-th frame of the sequence.
We implemented multiple deep architectures that learn to predict the set of events $Y_t$ that characterize frame $\mathcal{I}_t$, given the sequence $\mathcal{S}_t$. Formally, if $f(\cdot)$ is a generic deep neural network, we have
\[
\nonumber
Y_t = f(t | \mathcal{S}_{t}) \ ,
\label{pred}
\]
where $Y_t = \{y_{t,h},\ h = 1,\ldots,|\mathcal{E}| \}$ is a set of predictions of the considered events $\mathcal{E}$ (in this work, $|\mathcal{E}|=4$). In particular, $y_{t,h} \in \{0,1\}$, where $y_{t,h}=1$ means that the $h$-th event is predicted at time $t$.

% GENERIC STRUCTURE OF THE GENERIC NET
Before being processed by the network, each frame $\mathcal{I}_i$ is converted into one of the three representations that we described in Section~\ref{sec:data}, generically indicated here with $r_i$,
\begin{equation}
\nonumber
r_i = \texttt{frame\_representation}(\mathcal{I}_i) \ .
\label{rep}
\end{equation}
Our deep architecture $f(\cdot)$ is then composed of four computational stages, and each of them projects its input into a new latent representation. Stages consist of a feature extraction module $\texttt{feature\_extraction}(\cdot)$, a sequence representation module $\texttt{sequence\_representation}(\cdot)$, a prediction layer $\texttt{predictor}(\cdot)$, and a decision function $\texttt{decision}(\cdot)$ that outputs $Y_t$, 
\begin{eqnarray}
\label{a} q_t &=& \texttt{feature\_extraction}(r_t)\\
\label{b} s_t &=& \texttt{sequence\_representation}(q_t,\ s_{t-1} )\\
\label{c} p_t &=& \texttt{predictor}(s_t) \\
\label{d} Y_t &=& \texttt{decision}(p_t) \ .
\label{f}
\end{eqnarray}
Eq. (\ref{a}) is responsible of extracting features from $r_t$, building a new representation $q_t$ of the current frame. We implemented multiple extractors, in function of the method selected to produce $r_t$ (we postpone their description).
Eq. (\ref{b}) encodes the sequence of frames observed so far. The sequence representation $s_t$ is computed by updating the previous representation $s_{t-1}$ with the current input $r_t$. This is implemented with a Recurrent Neural Network (RNN), where $s$ is the hidden state of the RNN. In particular, we used a Long Short Term Memory RNN (LSTM) \cite{lstms}, and we also experienced multiple layers of recurrence (2 layers). Eq. (\ref{c}) is a fully connected layer with sigmoidal activation units, that computes the event prediction scores $p_t\in[0,1]^{|\mathcal{E}|}$. We indicate with $p_{t,h}$ the $h$-th component of $p_{t}$, and Eq. (\ref{d}) converts it into the binary decision $y_{t,h}$. We implemented each decision $y_{t,h}$ to be the outcome of a thresholding operation on $p_{t,h}$, so that
\[
\nonumber
y_{t,h} = \left\{\begin{array}{@{}ll@{}}
      1, & \mbox{if}\ p_{t,h} \geq \gamma_{h} \\
      0, & \mbox{otherwise}
    \end{array}\right.
\]
where $\gamma_{h}\in(0,1)$ is the threshold associated to the $h$-th event.

% LOSS FUNCTION AND TRAINING PROCEDURE
We are given a training set composed of fully labeled video clips, so that we have a ground truth label $\hat{Y}_t=\{\hat{y}_{t,h}, h=1,\ldots,|\mathcal{E}| \}$ on each frame. For each sequence, the time index $t$ spans from $1$ to the length of the sequence itself, and we set $s_{0}$ to be a vector of zeros. We trained our network by computing a loss function that, at each time instant, consists of the cross-entropy between the event-related output values and the ground truth,
\[
\nonumber
\mathcal{L}_t = \sum_{h=1}^{|\mathcal{E}|} \left\{ w_{h} \cdot [ -\hat{y}_{t,h} \cdot \log(p_{t,h})] - (1-\hat{y}_{t,h}) \cdot \log(1 - p_{t,h}) \right\} \ .
\label{cross}
\]
Notice that we introduced the scalar $w_{h} > 0$ to weigh the contribute of the positive examples of class $h$. As a matter of fact, it is crucial to give larger weight to those events that are rarely represented in the training data, and our experience with the data of Section \ref{sec:data} suggests that an even weighing scheme frequently leads to not promising results (we choose $w_{1}=10$, $w_{2}=40$, $w_{3}=30$, $w_{4}=100$, following the event ordering of Section \ref{sec:problem}).

\begin{figure}[!ht]
\centering
\includegraphics[width=0.70\textwidth,trim={0.5cm 1.0cm 4.3cm 0cm},clip]{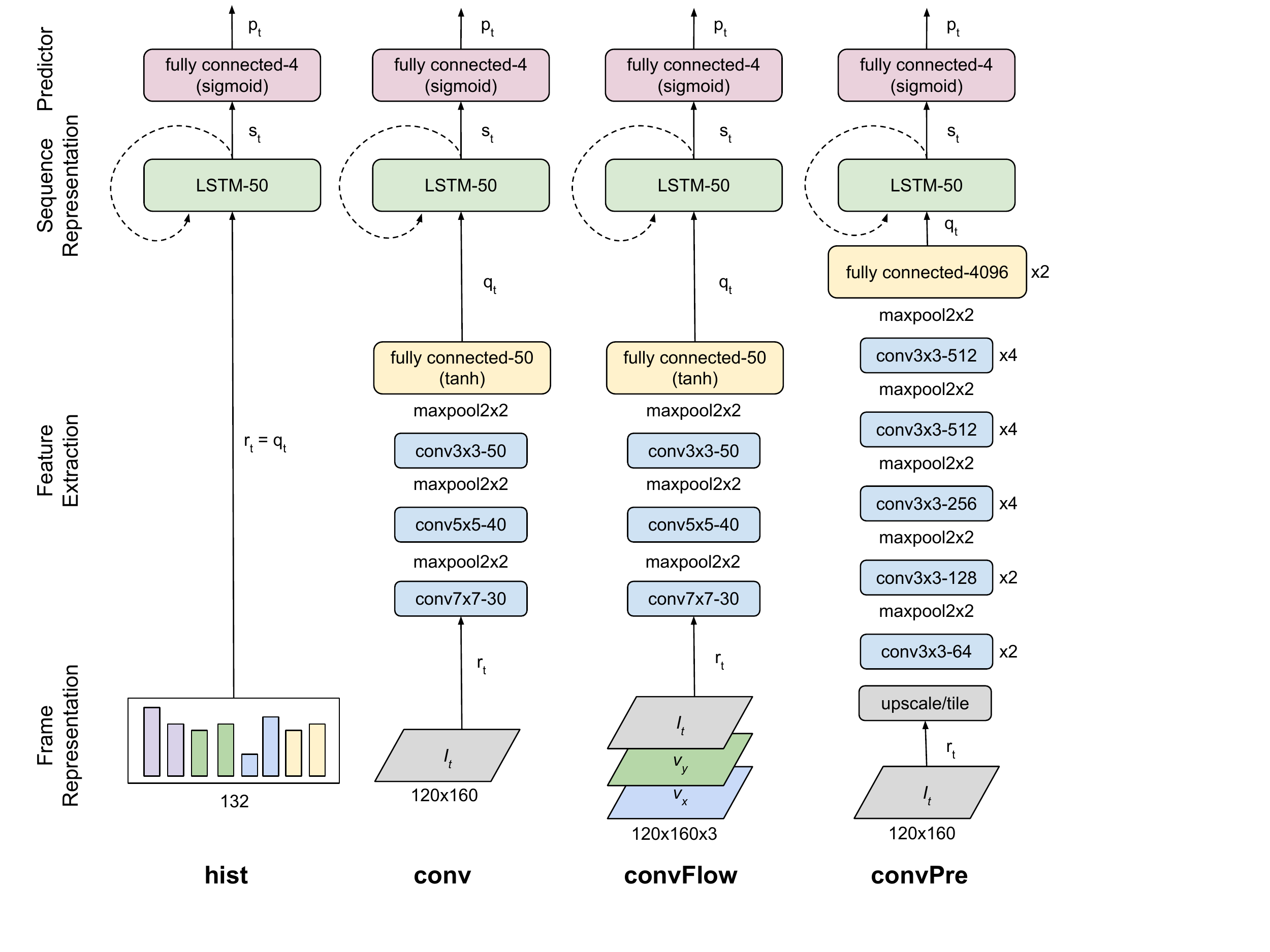}
\caption{Deep architectures applied to our task. Layer names are followed by the suffix \textit{-n}, where $n$ is the number of output units (or the size of the hidden state in LSTM).
We use the ReLu activation, unless differently indicated in brackets. In convolutional and pooling layers we report the size of their spatial coverage (e.g., $k \times k$). In Section \ref{sec:experiments} we evaluate several variants of these nets.}
\label{fig:nets}
\end{figure}

% SPECIFIC NETS
We evaluated four different deep networks that follow the aforementioned computations, and that are depicted in Fig.~\ref{fig:nets}, together with several numerical details. The networks differ in the frame representation $r_t$ that they process (Eq. (\ref{rep})) and in the way they implement the \texttt{feature\_extraction} function of Eq. (\ref{a}). The first network, referred to as \textit{hist}, processes the histogram of the motion features in the input frame, that are fed to the RNN without further processing ($q_t = r_t$). The second network, \textit{conv}, is based on the appearance-only representation of each frame, i.e. $r_t = \texttt{gray}(\mathcal{I}_t)$, and it extracts features using a Convolutional Neural Network (CNN) with 3 layers (we also tested configurations with 2 layers). When the frame representation consists of the appearance $\texttt{gray}(\mathcal{I}_t)$ paired with the motion field $(v_x, v_y)$, then \textit{conv} becomes the \textit{convFlow} network. Finally, we also considered the effects of transfer learning in the \textit{convPre} model, where we modified the \textit{conv} net by plugging a pre-trained VGG-19 convolutional network \cite{simonyan2014very} in Eq. (\ref{a}). VGG-19 is composed of 19 layers and trained using the ImageNet database, so $r_t$ is first rescaled/tiled to $224 \times 224 \times 3$ to match the size of the ImageNet data. 

% FIGURE (3 COLUMNS)

%% file: experiments.tex
% !TEX root = icann2018.tex

%We computed the three different representations described in Section \ref{sec:data}. In summary, we extracted from each video an hand engineered (through histograms) motion based feature representation, an appearance based representation (frames) and a mixed appearance-motion representation.
We divided our dataset sets of video clips into three groups for fitting, validating and testing our models, with a ratio of  $70\%, 20\%, 10\%$, keeping the original distribution of events in each split.
We selected the F1 measure to evaluate the models of Section \ref{sec:model}, and since some events occur very rarely in the data (see Section \ref{sec:data}), we computed the F1 for each single event class.
In particular, for every tested architecture, we selected the optimal value of the decision threshold $\gamma_{h}$ ensuring the best performances on the validation set (testing multiple values in $[0.1,0.9]$). We trained our networks with stochastic gradient-based updates that occur after having processed each video clip, and we used the Adam optimizer with a learning rate of $3\cdot10^{-5}$, processing the training data for $350$ epochs.
The training times are reported in Tab. \ref{tab:tabtempi}, considering a system equipped with %an Intel \textregistered Core\textsuperscript{TM} i7-3970X CPU @ 3.50GHz and 
an NVidia GTX Titan GPU (recall that the CNN of \textit{convPre} is pre-trained).

%\begin{table}[]
%\centering
%\caption{Average training times of the compared models.}
%\label{tab:tabtempi}
%\begin{tabular}{@{}lll@{}}
%\toprule
%Model  &   \multicolumn{2}{l}{ Training time (hours)} \\ \midrule
% &     1 layer RNN      &  2 layer RNN        \\
% \midrule
%Conv &      15.12     &      22.44     \\
% ConFlow&     17.05      &      21.96    \\
%hist &      2.03     &    3.06      \\
% convPre &     3.91      &   6.41       \\ \bottomrule
%\end{tabular}
%\end{table}
\begin{table}[]
\centering
\caption{Avg training times (hours). Frame representations were precomputed.}
\vskip 1mm
\label{tab:tabtempi}
\begin{tabular}{l|cccc}
\hline
\hline
 & $\ $$\ $hist$\ $$\ $ & $\ $conv$\ $ & convFlow & convPre\\
 \hline
1 Layer RNN$\ $ & $2.03$ & $15.12$ & $17.05$ & $3.91$ \\
2 Layers RNN$\ $ & $3.06$ & $21.96$ & $22.44$ & $6.41$ \\
\hline
\hline
\end{tabular}
\end{table}

\begin{figure}
\centering
\includegraphics[width=0.8\textwidth]{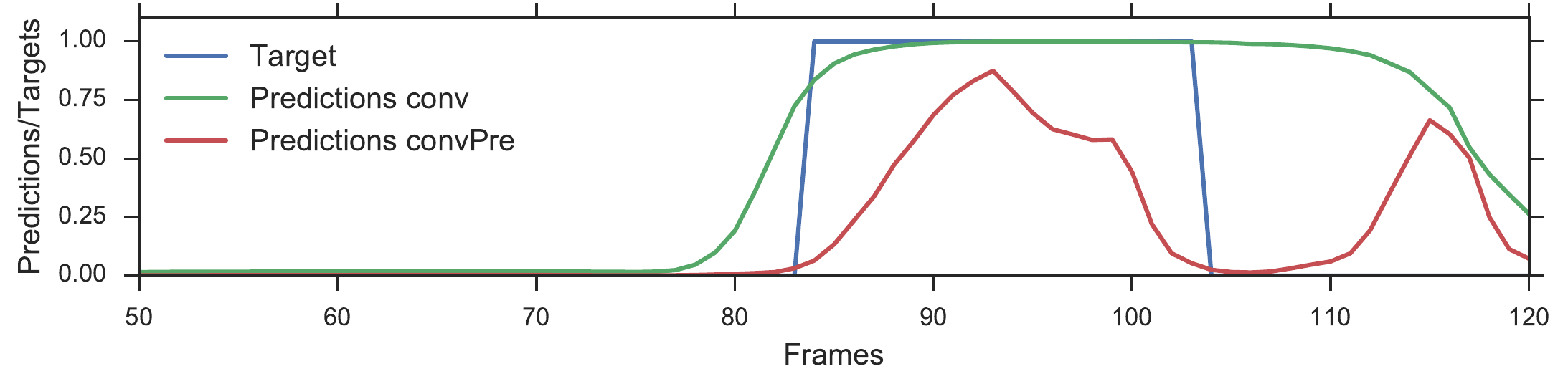}
\caption{Comparing predictions and ground truth in a  ``departing'' event.}
\label{fig:qualitative}
\end{figure}

We summarize the best performances obtained, for each class of event, by the \textit{hist, conv, convFlow, convPre} models of Fig. \ref{fig:nets} with multiple layers or recurrence. Depending on the model, we also evaluated different sizes of the recurrent state dimension $h$ ($20,50,132$ for \textit{hist}, $30,50,200$ for \textit{convPre}), or number of convolutional layers $\ell$ (2 or 3 layers for both \textit{conv} and \textit{convFlow}). 
%In particular, we report the $F1$ measure obtained by the \textit{hist} model with three different values of the recurrent state dimension $h_t$ of the LSTM network ($20, 50, 132)$.
%Regarding the \textit{conv} and \textit{convFlow} models, we report values obtained with architectures composed by 2 and 3 convolutional layers (respectively composed by 30 -- 40 and 30 -- 40 -- 50 feature maps).  
%In the \textit{convPre} model we indicate values obtained varying the state size dimension. ( $30, 50, 200 $).
% \usepackage{booktabs}
\begin{table} [h!]
\centering
%LCRCR
\caption{Performances (F1) of the compared models.}\label{beta}
\vskip 1mm
\begin{tabular}{LLCCCCCCCCCC}
\hline
\hline
\multicolumn{1}{l}{ } & & 
\multicolumn{3}{c}{{hist}}    &
\multicolumn{2}{c}{{conv}}  &
\multicolumn{2}{c}{{convFlow}} &
\multicolumn{3}{c}{{convPre}}  \\ 
\cmidrule(lr){3-5}
\cmidrule(lr){6-7}
\cmidrule(lr){8-9}
\cmidrule(lr){10-12}

\multicolumn{1}{c}{}& & 
\multicolumn{1}{c}{$h=20$} &
\multicolumn{1}{c}{$50$}     &
\multicolumn{1}{c}{$132$} &
\multicolumn{1}{c}{$\ \ell =2$}    &
\multicolumn{1}{c}{$3\ $}  &
\multicolumn{1}{c}{$\ \ell =2$}    &
\multicolumn{1}{c}{$3\ $}  &
\multicolumn{1}{c}{$h=30$} &
\multicolumn{1}{c}{$50$}     &
\multicolumn{1}{c}{$200$}   \\

\hline& & & & & & & & & & \\
[-2.5mm]\multirow{2}{*}{\footnotesize(stationary)$\ $} & \mbox{1\ Layer\ RNN} & 0.63  & 0.89 & 0.86 & \textbf{0.98} & 0.94 & 0.91 & 0.79 & 0.95 & 0.92 & 0.90\\
 & \mbox{2\ Layers\ RNN} & 0.87 & 0.85 & 0.87 &  0.87 & 0.91 &  0.76 & 0.82 & 0.95 & 0.91 & 0.89 \\
[1mm]\hline& & & & & & & & & & \\
[-2.5mm]\multirow{2}{*}{\footnotesize(departing)$\ $} & \mbox{1\ Layer\ RNN} & 0.56  & 0.63 & 0.73 & 0.61 & 0.79 & 0.63 & 0.57 & 0.59 & 0.63 & 0.58 \\
 & \mbox{2\ Layers\ RNN} & 0.48 & 0.66 & 0.72 &  0.74 & \textbf{0.82} & 0.52 & 0.63 & 0.74 & 0.81 & 0.68 \\
[1mm]\hline& & & & & & & & & & \\
[-2.5mm]\multirow{2}{*}{\footnotesize(wrong--way)$\ $} & \mbox{1\ Layer\ RNN} & 0.59  & 0.84 & 0.89 & 0.92 & 0.93 & 0.89 & 0.92 & \textbf{0.96} & 0.93 & 0.94 \\
& \mbox{2\ Layers\ RNN} & 0.83 & 0.87 & 0.89 &  0.86 & 0.88 & 0.86 & 0.88 & 0.94 & 0.93 & 0.91 \\
[1mm]\hline& & & & & & & & & & \\
[-2.5mm]\multirow{2}{*}{\footnotesize(car crash)$\ $} & \mbox{1\ Layer\ RNN} & 0.65  & 0.61 & 0.70 & 0.85 & 0.64 & 0.79 & 0.78 & 0.80 & 0.78 & 0.77 \\
& \mbox{2\ Layers\ RNN} & 0.56 & 0.33 & 0.91 &  0.75 & 0.50 & 0.74 & 0.74 & \textbf{0.95} & 0.86 & 0.90 \\
[1mm]\hline& & & & & & & & & & \\
[-2.5mm]\multirow{2}{*}{\footnotesize(\textit{average})$\ $} & \mbox{1\ Layer\ RNN} & \footnotesize 0.61  & 0.74 & 0.80 & 0.84 & 0.83 & 0.81 & 0.77 & 0.83 & 0.82 & 0.80 \\
& \mbox{2\ Layers\ RNN} & 0.69 & 0.68 & 0.85 &  0.81 & 0.78 & 0.72 & 0.77 & \textbf{0.90} & 0.88 & 0.85 \\
[1mm]\hline
\hline
\end{tabular}

\end{table}
These results show that the \textit{hist} approach generally performs worse than the other models, and that convolutional architectures are a better solution to the proposed task, by virtue of their capability to extract autonomously relevant representation from images. When using 2 layers of RNNs, the configuration of  \textit{hist} with $h=132$ leads to more competitive results, that, however, are paired with a larger computational burden than the CNN-based models due to the cost of computing its hand-engineered features. The \textit{conv} model with only two convolutional layers shows good results paired with a computational cost that can be tolerated in real-time applications. The addition of the motion related information (\textit{convFlow}) does not seem to help the performances. This can be explained by the fact that an architecture composed by a combination of a CNN together with a RNN is able by itself to grasp the temporal dynamics of a video, making an addition of optical flow features worthless. The use of a pre-trained network (\textit{convPre}) leads to the best performances, on average, even if with a more costly inferential process. Finally, we notice that using 2 layers of RNNs does not add useful information to the \textit{conv} model, while it always helps in \textit{convPre}, mostly due to larger number of high--level features that are extracted by the CNN, where the system seems to find longer-term regularities (more easily captured by multiple layers of recurrence). The event class where all the models have shown worse performances is ``departing'', that we explain by the larger incoherence in the training data in defining the beginning and, mostly, the ending frames of the event. In Fig. \ref{fig:qualitative} we report an example that compares a prediction and the ground truth (test set), showing the mismatch in the ending-part of the event.
% TODO
% per ogni classe mettere in BOLD il migliore
%1 HIST peggiore degli altri in generale, architetture convoluzionali preferibili!
%2 CONV + flusso ottico non sembra essere utile ai fini di migliorare le performance, perch� la rete ricorrente � gi� in grado di estrarre le informazioni utili a catturare variazioni temporali
%3 PREtrained feature sembrano aiutare in media
%4 molteplici layer ricorrenti non sembrano aggiungere informazione utile
%5 conv 2 layers � BUONA e anche convPre � BUONA (quindi si vede che transfer learning � buono e rende gli addestramenti molto veloci facendo riferimento alla tabella dei tempi di training) => BEST � conv2 layers perch� pi� efficiente (modello pi� piccolo) e vogliamo modelli real-time

\vskip-3mm
% MANCA
%1 pesi loss
%2 esperimento qualitativo DEPARTING (si dice che � difficile predirre dove inizia o finisce l'evento perch� � molto soggettivo e il dataset � piccolo)

%% file: conclusions.tex
% !TEX root = icann2018.tex
\vskip-3mm
We described a deep-network-based implementation of an ATMS (Advanced Traffic Management System) that predicts a set of events while processing videos of traffic on highways. We performed a detailed analysis of a real-world video data collection, investigating four classes of traffic events. We reported the results of an experimental evaluation that involved multiple representations of the input data and different deep architectures composed of a stack of convolutional and recurrent networks. Our results have shown that these networks can efficiently learn the temporal information from the video stream, simplifying the feature engineering process and making very promising predictions. We also proved the benefits of transferring the representations learned on a generic image classification task. Our best architectures will be tested online in real operative conditions.
\vskip-5mm